\documentclass[letterpaper, 10 pt, conference]{ieeeconf}  

\IEEEoverridecommandlockouts                              

\usepackage{graphics} 
\usepackage{epsfig} 
\usepackage{mathptmx} 
\usepackage{times} 
\usepackage{amsmath} 
\usepackage{amssymb}  
\usepackage{graphicx}
\usepackage[OT1]{fontenc}
\usepackage{multirow}
\usepackage{booktabs}
\usepackage{tikz}
\usepackage{stfloats}
\usepackage{array}
\usepackage{pifont}
\newcommand{\cmark}{\ding{51}}%
\newcommand{\xmark}{\ding{55}}%
\usepackage{threeparttable}
\usepackage{cite}
\newcommand\shift{-3.7mm}

\usepackage[colorlinks=true, linkcolor=black, urlcolor=black]{hyperref}
\usepackage[dvipsnames]{xcolor}

\title{\LARGE \bf
Two Degree-of-Freedom Vibratory Transport in a Grasp}

\author{
\thanks{}%
}

\author{C. L. Yako, Shenli Yuan, and Kenneth Salisbury
\thanks{Connor Yako and Shenli Yuan are with the Department of Mechanical Engineering, Stanford University, Stanford, CA 94305 USA (email: \{connor.yako, shenliy\}@stanford.edu). Kenneth Salisbury is with Department of Computer Science, Stanford University, Stanford, CA 94305 USA (email: kenneth.salisbury@gmail.com).}%
}

\begin{document}

\maketitle
\thispagestyle{empty}
\pagestyle{empty}

\begin{abstract}

In this paper, we use asymmetric vibrations to demonstrate two degree-of-freedom (DoF) in-hand manipulation of grasped parts. 
The asymmetric vibrations are achieved through closed-loop position control of a moving surface, which applies a periodic stick-slip waveform to the part to be manipulated.
We show analytically how two vibratory waveform parameters, the sticking acceleration and the slipping acceleration, affect average part velocity when moving against gravity.
The theoretical trends are then validated using an experimental setup where the squeeze force is controlled and part motion is recorded by a high-resolution encoder.
We also develop a 2-DoF vibratory surface capable of translation in one direction and rotation about the surface normal.
Using two of these 2-DoF surfaces in a parallel jaw gripper configuration, we bidirectionally translate and rotate a variety of grasped parts, as well as demonstrate that the same waveform trends for translation also persist for in-plane rotation.

\end{abstract}

\section{INTRODUCTION}\label{section: introduction}

{\color{black}There has been an effort toward non-anthropomorphic hand designs able to perform complex manipulations without significant control overhead or overly complicated sensing. 
The Roller Graspers~\cite{yuan2022robot} are a prime example of this, which used controllable rollers at the fingertips, termed ``active surfaces,'' to drive a grasped part to a goal pose. 
The purpose of this work is to continue in that same vein of active surfaces for manipulation~\cite{datseris1985principles,tincani2012velvet,ma2016manipulation,gomezdegabriel2021adaptive,cai2023hand,xie2024belted}, but instead of rollers or belts, we utilize a vibrating surface. 
A vibrating surface potentially offers several benefits over other active surfaces.
In theory, a fully planar vibrating surface can achieve holonomic part motion. 
In contrast, rolling fingertips are non-holonomic, since their rollers must reorient to drive a part in a desired direction. 
A vibrating surface also provides a larger contact area, yielding a more stable grasp than rollers with point or line contact. 
A device akin to an omni-conveyor~\cite{keek2021design} could offer similar grasp stability, but a 3-DoF vibrating surface is arguably mechanically simpler (e.g., a parallel 3-$RRR$ mechanism).
These potential advantages motivate our investigation into the practicality of vibration-based manipulation.
%
}

The use of vibrations as a means of manipulating parts has a long history, particularly in non-prehensile transport.
Typical vibration-based manipulation strategies leverage \textit{time asymmetry}, where a drive surface spends a majority of the cycle time moving in the desired transport direction.
Reznik and Canny noted this strategy in their ``Coulomb Pump'' part feeder, which assumed continuous sliding in a bang-coast-bang waveform to propel parts~\cite{reznik1998coulomb}. 
Quaid removed the always sliding assumption and instead used a low acceleration \textit{sticking} phase followed by a high acceleration \textit{slipping} phase~\cite{quaid1999feeder}.
Umbanhowar and Lynch extended Quaid's work by adding a third phase where the surface accelerates up to the part, maximizing cycle averaged part velocity~\cite{umbanhowar2008optimal}.
They also removed the 1D oscillation assumption and showed that transverse vibrations alter the effective gravity, further increasing cycle averaged part velocity. 

Recent work by Nahum and Sintov cleverly implemented this idea by using an eccentric rotating mass (ERM) motor~\cite{nahum2022robotic}.
Rotation of the motor modulated both the normal and friction forces, facilitating easier transitions between sticking and slipping and strengthening the sticking phase.  
Experiments showed that the device could drive part translation against gravity for inclinations up to $60^\circ$ from horizontal (part velocity around 2 $\text{mm/s}$ at $60^\circ$), though stronger actuation was required beyond that angle.
A follow up work by Binyamin et al. added orientation control to this device by leveraging part rotation when the center-of-mass (CoM) was not along the line connecting the two contact points~\cite{binyamin2024vibration}.
\begin{figure}[t]
    \centering
    \includegraphics[width=0.43\textwidth]{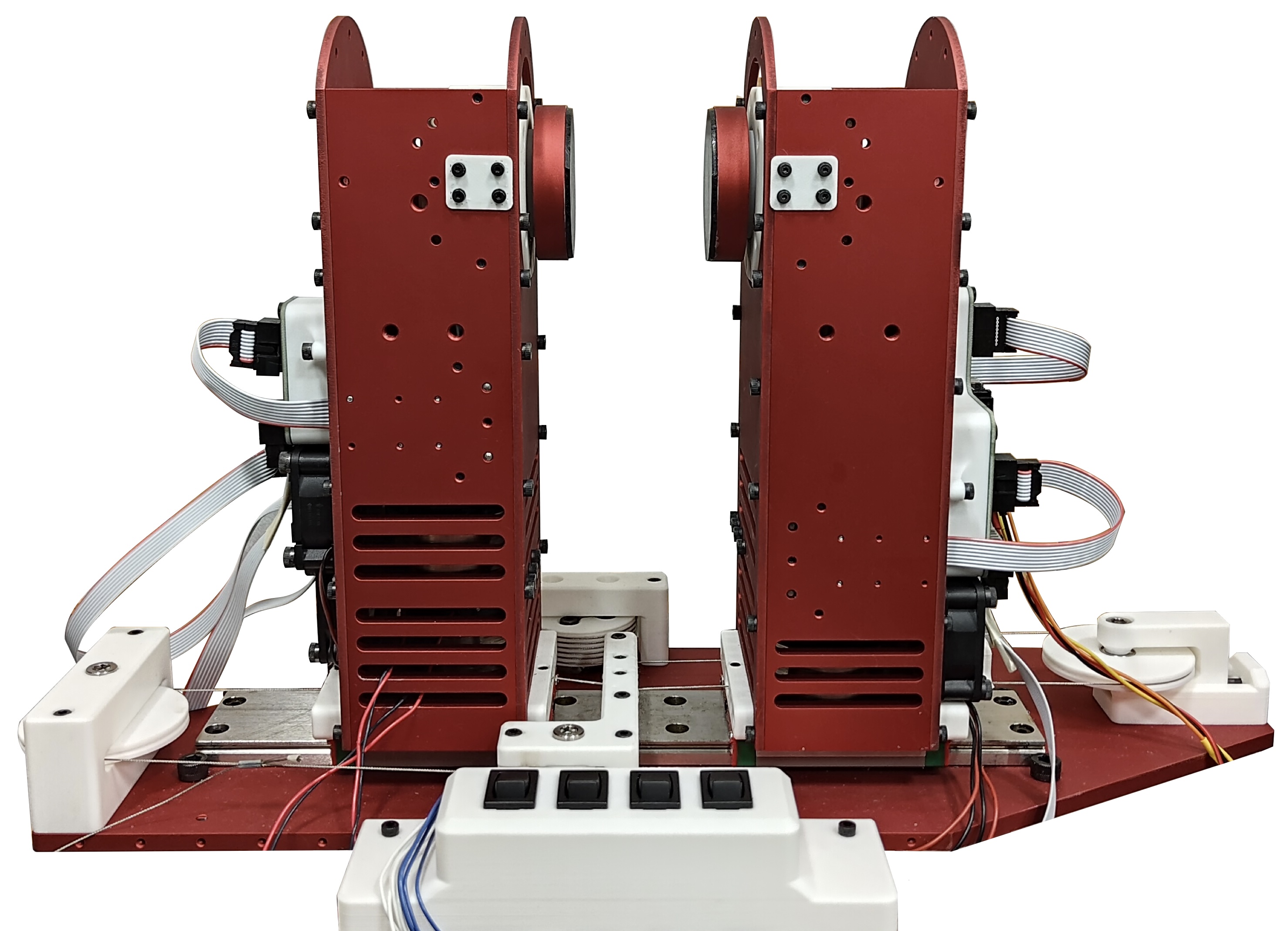}
    \caption{Parallel jaw gripper with 2-DoF vibrating surfaces in each finger. 
    Rocker switches are used to control on/off, translation/rotation transport modes, and up/down or clockwise/counterclockwise directions.
    }
    \label{figure: cover photo}
    \vspace{\shift}
\end{figure}

Our previous work significantly improved the maximum angle of transport by using voice-coil based impact motors---typically used for haptic applications---to drive surface motion~\cite{yako2024vertical}.
These impact-induced accelerations were large enough to instigate slip, and a variety of parts were shown to be transported directly against gravity. 
However, because of the unpredictable dynamics associated with the impact motors a human-in-the-loop was required to tune the shape of the current waveform, as well as adjust the squeeze force until motion was observed.
%

In this paper, we extend our prior vertical transport work by developing a similarly capable asymmetric vibration device that replaces unpredictable impact motors with controllable voice coil actuators (VCAs). 
These VCAs allow us to systematically characterize how vibration waveform parameters influence part motion, while enabling higher-DoF vibrating surfaces.
The paper proceeds as follows. Section~\ref{section: dynamics} presents the system dynamics for vertical transport, derives the average velocity of a part moving against gravity under a stick–slip surface waveform, and examines how velocity varies with waveform acceleration parameters. 
Section~\ref{section: device design} describes the position-controlled vibration system and its implementation as a 2-DoF vibrating finger. 
Section~\ref{section: experimental setup} details the experimental setup used to validate the velocity trends, and Section~\ref{section: results} discusses the results. 
In Section~\ref{section: gripper}, we introduce a parallel gripper with a 2-DoF vibratory surface in each finger, demonstrating bidirectional translation and rotation of various parts. 
Section~\ref{section: discussion} concludes with an evaluation of gripper performance and practical design considerations.
{\color{black}Our main contributions are: (1)~demonstrating vertical vibratory transport and part rotation in the presence of gravity without relying on impact-based accelerations, and (2)~characterizing and experimentally validating how waveform parameters affect linear and angular part velocity.}

\section{DYNAMICS}\label{section: dynamics}



\subsection{Dynamical Model}

\begin{figure}[t]
    \centering
    \includegraphics[width=0.44\textwidth]{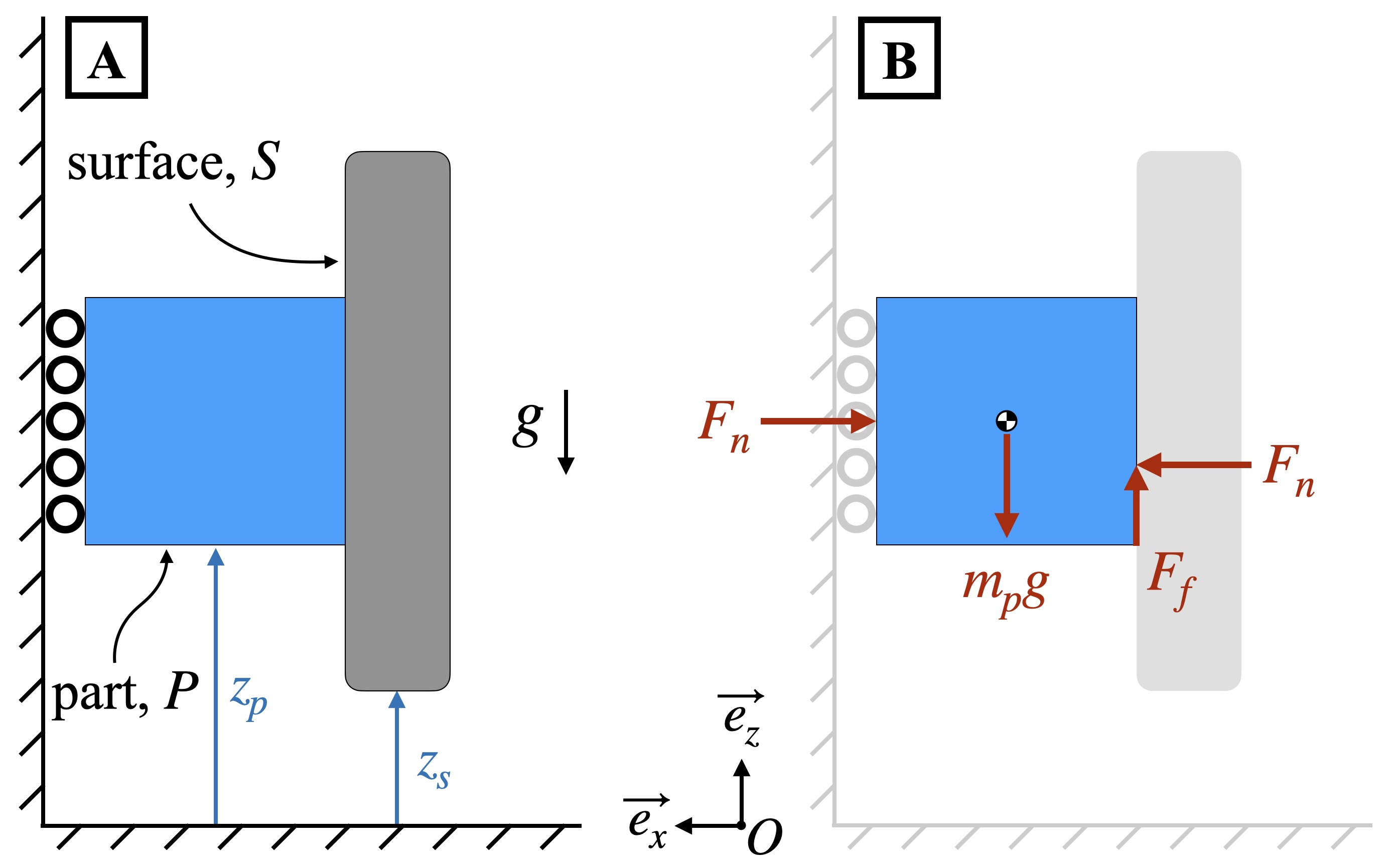}
    \caption{Kinematics (A) and dynamics (B).
    The part, $P$, and the surface, $S$ are constrained to move purely along the $\Vec{e_z}$ direction.
    The rollers on the left side of $P$ indicate sliding on a frictionless surface.
    This figure was reproduced from our previous paper~\cite{yako2024vertical}.
    }
    \label{figure: kinematics and dynamics}
    \vspace{\shift}
\end{figure}

The system is shown in Fig.~\ref{figure: kinematics and dynamics}. 
{\color{black}
Note that while only a single vibrating surface is modeled, this is valid for a physical system with any number of \textit{synchronized} vibrating surfaces contacting a part (see our previous work~\cite{yako2024vertical}), an assumption we make here.
}
The part $P$, with mass $m_P$, is subjected to the gravitational force $-m_P g \vec{e}_z$. 
It is laterally constrained between a vertically vibrating surface $S$---which applies a normal force $F_n \vec{e}_x$ and a tangential frictional force $F_f \vec{e}_z$---and an opposing vertical support represented by rollers that is assumed frictionless.
The interaction between $P$ and $S$ is modeled as dry Coulomb friction with static and kinetic coefficients $\mu_s \geq \mu_k$.
The positions of the surface and part relative to a fixed origin $O$ are denoted $z_S$ and $z_P$, respectively.  

The equation of motion of the part dotted with $\Vec{e}_z$ is: 

\begin{equation}\label{eq: EOM part}
    m_P \ddot{z}_P = F_f - m_P g
\end{equation}

When the part sticks to the surface, the velocities are equal ($\dot{z}_P = \dot{z}_S$) provided that the required frictional force $|F_f| = m_P|\ddot{z}_S + g|$ does not exceed $\mu_s F_n$. 
During ongoing sliding ($\dot{z}_P \neq \dot{z}_S$), the frictional force becomes  
$F_f = \mu_k F_n \,\text{sgn}(\dot{z}_S - \dot{z}_P)$,  
until the sticking conditions are restored.  
Given all of this, the equations of motion are:
\begin{flalign}
    \label{eq: EOM sticking}
    \quad \textit{sticking:} \quad & \dot{z}_P = \dot{z}_S, \quad -\frac{\mu_s F_n}{m_P} -g \leq \ddot{z}_S \leq \frac{\mu_s F_n}{m_P} - g&&\\
    \label{eq: EOM slipping}
    \quad \textit{slipping:} \quad &
    \dot{z}_P \neq \dot{z}_S, \quad \ddot{z}_P = \frac{\mu_k F_n}{m_P} \text{sgn}(\dot{z}_S - \dot{z}_P) - g&&
\end{flalign}
\subsection{Quaid Waveform}\label{section: quaid waveform}
We used the alternate sticking-slipping waveform from Quaid and adapted his expressions for part velocity for vertical transport.
In contrast to the velocity-maximizing three-phase motions of Umbanhowar and Lynch~\cite{umbanhowar2008optimal} and our previous work~\cite{yako2024vertical}, which rely on detailed knowledge of friction coefficients and part mass for precise timing of the waveform phases, Quaid’s method is independent of such parameters~\cite{quaid1999feeder}.
The profile consists of only two phases: a sticking phase where the part and surface move together, and a slipping phase in which the surface rapidly accelerates downward relative to the part.

The acceleration profile for the moving surface is equivalent to that defined by Quaid~\cite{quaid1999feeder} for horizontal transport and is defined by the sticking acceleration $a_s$, the maximum acceleration magnitude $a_{max}$, and the oscillation frequency $f = 1/T$:
%
%
\begin{equation}
\label{eq: quaid waveform}
\ddot{z}_S =
\begin{cases}
    a_{s}, & 0 \leq t \leq t_1, \\
    -a_{max}, & t_1 < t \leq T - t_1, \\
    a_{s}, & T - t_1 < t \leq T,
\end{cases}
\end{equation}
where $t$ denotes time within one period, $T$. 
The duration of the sticking intervals is chosen so that the surface has zero net displacement over one cycle:
\begin{equation}
\label{eq: quaid acceleration time}
    t_1 = \frac{T \cdot a_{max}}{2(a_s + a_{max})}
\end{equation}
The corresponding acceleration of the part is the following:
\begin{equation}
\ddot{z}_P =
\begin{cases}
    a_{s}, & 0 \leq t \leq t_1, \\
    -a_k, & t_1 < t \leq t_2, \\
    a_{s}, & t_2 < t \leq T,
\end{cases}
\end{equation}
where $a_k = \frac{\mu_k F_n}{m_P} + g$ and $t_2 = t_1 + \frac{a_s}{a_s + a_k}T$.
Although suboptimal in terms of average velocity, the Quaid waveform provides a practical alternative when system parameters such as $\mu_s$, $\mu_k$, or $m_P$ are uncertain or variable. 

\subsection{Effect of Surface Accelerations on Average Part Velocity}
The accelerations $a_s$ and $a_{max}$ govern how momentum is transferred from the vibrating surface to the part. 
For sustained sticking we have:
\begin{equation}
    \label{eq: sticking criterion}
    0 <\: a_s \:\leq \min\left(\frac{\mu_s F_n}{m_P} - g,\:a_s \text{ s.t. } A(a_s, a_{max}, T) \leq A_{max}\right)\\ 
\end{equation}
where $A_{max}$ is the maximum stroke of the system and $A(a_s, a_{max}, T)$ is the stroke for a given waveform:
\begin{equation}
    \label{eq: waveform amplitude}
    A(a_s, a_{max}, T) = \frac{a_{max} a_s}{8\left(a_{max} + a_s\right)}
T^2\end{equation}
For reliable initiation of slipping, $a_{max}$ must satisfy the following bounds:
\begin{equation}
    \label{eq: slipping criterion}
    \frac{\mu_s F_n}{m_P} + g <\: a_{max} \:\leq a_{system,max}
\end{equation}
where $a_{system,max}$ is defined by the system's maximum achievable acceleration.

Given \eqref{eq: EOM sticking} -- \eqref{eq: slipping criterion}, the average part velocity over one period, $v_{avg}$, is given as the following:

\begin{equation}
\label{eq: average part velocity}
    v_{avg} = \frac{a_s}{2f}\left(\frac{1}{1 + \frac{a_s}{a_{max}}} - \frac{1}{1 + \frac{a_s}{a_k}}\right)
\end{equation}
The effect of each surface acceleration on $v_{avg}$ can be seen by taking the partial derivatives:

\begin{align}
    \label{eq: a_max partial}
    \frac{\partial v_{avg}}{\partial a_{max}} &= \frac{a_s^2}{2f \left(a_{max} + a_s\right)^2}\\
    \label{eq: a_s partial}
    \frac{\partial v_{avg}}{\partial a_s} &= \frac{a_s\left(a_{max} - a_k\right)\left(a_s\left(a_{max} + a_k\right) + 2a_ka_{max}\right)}{2f \left(a_k + a_s\right)^2\left(a_{max} + a_s\right)^2}
\end{align}

Equation~\eqref{eq: a_max partial} shows that $v_{avg}$ grows monotonically with $a_{max}$. 
Physically, raising $a_{max}$ extends the sticking interval, allowing the part to be accelerated longer in the upward direction, while correspondingly reducing the duration of slip where deceleration occurs. 
With \eqref{eq: a_s partial}, the situation is more nuanced: although increasing $a_s$ reduces the sticking interval $t_1$, the stronger acceleration in that shorter window more than compensates. 
As a result, the net momentum transfer per cycle rises, provided $a_{max} > a_k$, which holds when $\mu_s \geq \mu_k$. 
%


\section{DEVICE DESIGN}\label{section: device design}

\begin{table}[t!]
\centering
\caption{Key System and Experimental Components}
\begin{tabular}{>{\centering\arraybackslash}m{0.4cm} >{\centering\arraybackslash}m{2.7cm} >{\centering\arraybackslash}m{1.5cm} >{\centering\arraybackslash}m{2.3cm}}
\toprule
\textbf{No.} & \textbf{Component} & 
\begin{tabular}[c]{@{}c@{}} \textbf{Manufacturer /} \\ \textbf{Supplier} \end{tabular} & 
\textbf{Part \#} \\
\midrule\midrule
\refstepcounter{enumi}\theenumi\label{comp: vca} & Voice Coil Actuator & BEI Kimco & LA18-18-000A\\\midrule
\refstepcounter{enumi}\theenumi\label{comp: vca bearing block} & Bearing Block (VCA) & Hiwin & MGN07CZ0H\\\midrule
\refstepcounter{enumi}\theenumi\label{comp: vca rail} & Rail (VCA) & Hiwin & MGNR07R\\\midrule
\refstepcounter{enumi}\theenumi\label{comp: motor driver} & Motor Driver & Pololu & 2993\\\midrule
\refstepcounter{enumi}\theenumi\label{comp: encoder} & Linear Encoder & Posic & ID4501L\\\midrule
\refstepcounter{enumi}\theenumi\label{comp: scale} & Encoder Scale & Posic & TPLS04-026\\\midrule
\refstepcounter{enumi}\theenumi\label{comp: experiment motor} & DC Motor & Hsiang Neng & HN-35GMB-1640Y\\\midrule
\refstepcounter{enumi}\theenumi\label{comp: loadcell} & Load Cell & Phidgets & FRC4163\_0\\\midrule
\refstepcounter{enumi}\theenumi\label{comp: polyurethane} & Polyurethane & McMaster & 8716K61\\\midrule
\refstepcounter{enumi}\theenumi\label{comp: gripper rail} & Rail (Finger) & Hiwin & MGWR15R\\\midrule
\refstepcounter{enumi}\theenumi\label{comp: gripper bearing block} & Bearing Block (Finger) & Hiwin & MGW15HZ0HM\\\midrule
\refstepcounter{enumi}\theenumi\label{comp: gripping motor} & Motor (Gripper) & Maxon & 148877\\
\bottomrule
\end{tabular}
\label{tab: system components}
\vspace{\shift}
\end{table}

We created a vibratory transport device that could generate sufficiently large accelerations to transport against gravity, while also being controllable in order to test the trends associated with $a_s$ and $a_{max}$ discussed in Section~\ref{section: dynamics}.
Note that the main system and experimental components are listed in Table~\ref{tab: system components} and are denoted with superscripts in the text. 
We selected a voice coil actuator$^{\ref{comp: vca}}$ (VCA) for its high peak acceleration---with a peak force of $48.93~\mathrm{N}$ and a moving coil mass of $0.0471~\mathrm{kg}$ the theoretical peak acceleration is roughly $100~g$.
For scale, the magnet is $44.45~\times~44.45~\times~19.05~\mathrm{mm}$.
To maintain this high acceleration, we attached the moving coil to a very-light-preload bearing block$^{\ref{comp: vca bearing block}}$ sliding on a lubricated rail$^{\ref{comp: vca rail}}$, which trades high rigidity for low-friction, and has a mass of 0.01 kg. 
%
The VCA was driven with a PWM-based motor driver$^{\ref{comp: motor driver}}$.
Position feedback was provided by non-contact linear encoders$^{\ref{comp: encoder}}$, with the scale$^{\ref{comp: scale}}$ mounted to the moving coil assembly and the sensor itself fixed to the aluminum frame.
The sensors were programmed to have a resolution of $5~\mu\mathrm{m}$.
The A/B pulses from the encoder have a typical output frequency of 1 MHz and were recorded by the hardware quadrature encoder pins on a Teensy 4.1 microcontroller.
A simple PD position control loop running at 40 kHz ensured accurate tracking of the Quaid waveform, which is a repeating quadratic with sharp peaks at $t = \frac{T}{2}$ when $a_{max}$ is large.
\begin{figure}[t!]
    \centering
    \includegraphics[width=0.48\textwidth]{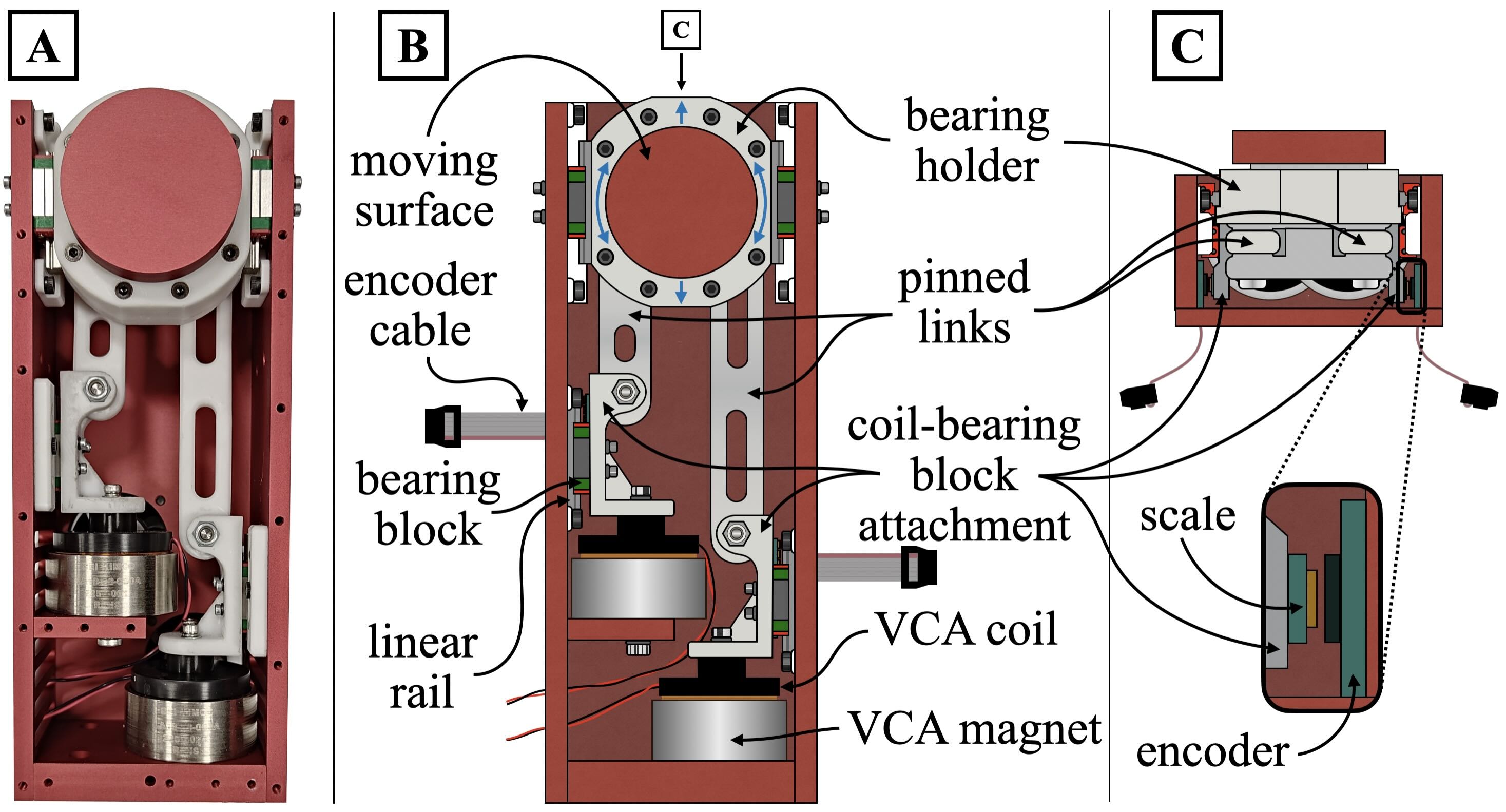}
    \caption{Design of a single finger.
    (A) Actual finger with the front cover removed.
    (B) Annotated graphic of the finger.
    Custom parts are shown as either red, white, or off-white, with the red parts anodized aluminum, the white parts are PLA, and the off-white parts are White V5 resin. 
    The two moving coils are linearly constrained by low-friction bearing blocks and rails.
    Their motion is transferred to the moving surface through two links which are pinned at either end.
    When the coils move synchronously, the bearing holder and surface move linearly.
    When the coils are 180$^\circ$ out-of-phase, the bearing holder is stationary and the surface rotates.
    (C) Front view of the finger. The encoder scale and sensor (differential transformer) are shown, with the scale glued to the coil-bearing block attachment.
    }
    \label{figure: finger design}
    \vspace{\shift}
\end{figure}

We took these components and integrated them into a finger design with a 2-DoF output surface, as shown in Fig.~\ref{figure: finger design}.
The moving coils transfer their linear motion to the output surface through two links, which are pinned at either end with titanium shoulder screws and ceramic ball-bearings to minimize mass and friction. 
The bearing holder is constrained to move linearly by a pair of bearing blocks$^{\ref{comp: vca bearing block}}$ and rails$^{\ref{comp: vca rail}}$.
When the coils move synchronously, the bearing holder and surface move linearly, and when they move 180$^\circ$  out-of-phase the bearing holder is stationary and the surface rotates.
%
%
Moving the coils in-phase but with differing amplitudes leads to a superposition of surface translation and rotation---this was not something we investigated in this work, but could be useful for closed-loop position control of a manipulated part.
Fans on the backside of the finger cooled the magnets and coils through cutouts in the frame.

\section{EXPERIMENTAL SETUP}\label{section: experimental setup}

\begin{table*}[ht!]
\centering
\caption{Experimental parameters and measured friction coefficients}
\begin{threeparttable}
\begin{tabular}{ccccccccccccccccc}
\toprule
\multirow{2}{*}{\textbf{Experiment \#}} & 
\multirow{2}{*}{\textbf{Fixed Parameters}} & 
\multirow{2}{*}{\textbf{Varied Parameter}} & 
\multicolumn{11}{c}{\textbf{Normal Force (N)}} & 
\multirow{2}{*}{\textbf{$\mu_s$}} & 
\multirow{2}{*}{\textbf{$\mu_k$}}\\ 
\cmidrule(lr){4-14}
 & & & 50 & 48 & 46 & 44 & 42 & 40 & 38 & 36 & 34 & 32 & 30 \\ 
\midrule\midrule
\textbf{1} & \multirow{4}{*}{$f = 20~\mathrm{Hz}$, $a_s = 0.7~g$} & $a_{max} = 20~g$  & \checkmark & \checkmark & \checkmark & \checkmark & \checkmark & \checkmark & \checkmark & \checkmark & \checkmark & \checkmark & \checkmark & 0.37 & 0.35 \\ 
\textbf{2} &  & $a_{max} = 15~g$ & \checkmark & \checkmark & \checkmark & \checkmark & \checkmark & \checkmark & \checkmark & \checkmark & \checkmark & \checkmark & \checkmark & 0.44 & 0.31 \\ 
\textbf{3} &  & $a_{max} = 10~g$ & \checkmark & \checkmark & \checkmark & \checkmark & \checkmark & \checkmark & \checkmark & \checkmark & \checkmark & \checkmark & \checkmark & 0.46 & 0.35\\ 
\textbf{4} &  & $a_{max} = 5~g$ & \checkmark & \checkmark & \checkmark & \checkmark & \checkmark & \checkmark & \checkmark & \checkmark & \checkmark & \checkmark & \checkmark & 0.49 & 0.30\\ 
\midrule
\textbf{5} & \multirow{4}{*}{$f = 15~\mathrm{Hz}$, $a_{max} = 10~g$} & $a_{s} = 0.4~g$ & \checkmark &  &  &  &  & \checkmark &  &  &  &  & \checkmark & 0.47 & 0.35\\ 
\textbf{6} &  & $a_{s} = 0.6~g$ & \checkmark &  &  &  &  & \checkmark &  &  &  &  & \checkmark & -- & --\\ 
\textbf{7} &  & $a_{s} = 0.8~g$ & \checkmark &  &  &  &  & \checkmark &  &  &  &  & \checkmark & -- & --\\
\textbf{8} &  & $a_{s} = 1.0~g$ & \checkmark &  &  &  &  & \checkmark &  &  &  &  & \checkmark & 0.42\tnote{\dag} & 0.39\tnote{\dag} \\ 
\bottomrule
\end{tabular}
\begin{tablenotes}
    \item[\dag] The friction coefficients for Experiment 8 were measured \textit{after} their corresponding experiment, while the rest were measured \textit{before}.
\end{tablenotes}
\end{threeparttable}
\label{tab: experimental parameters}
\vspace{\shift}
\end{table*}

\begin{figure}[t]
    \centering
    \includegraphics[width=0.48\textwidth]{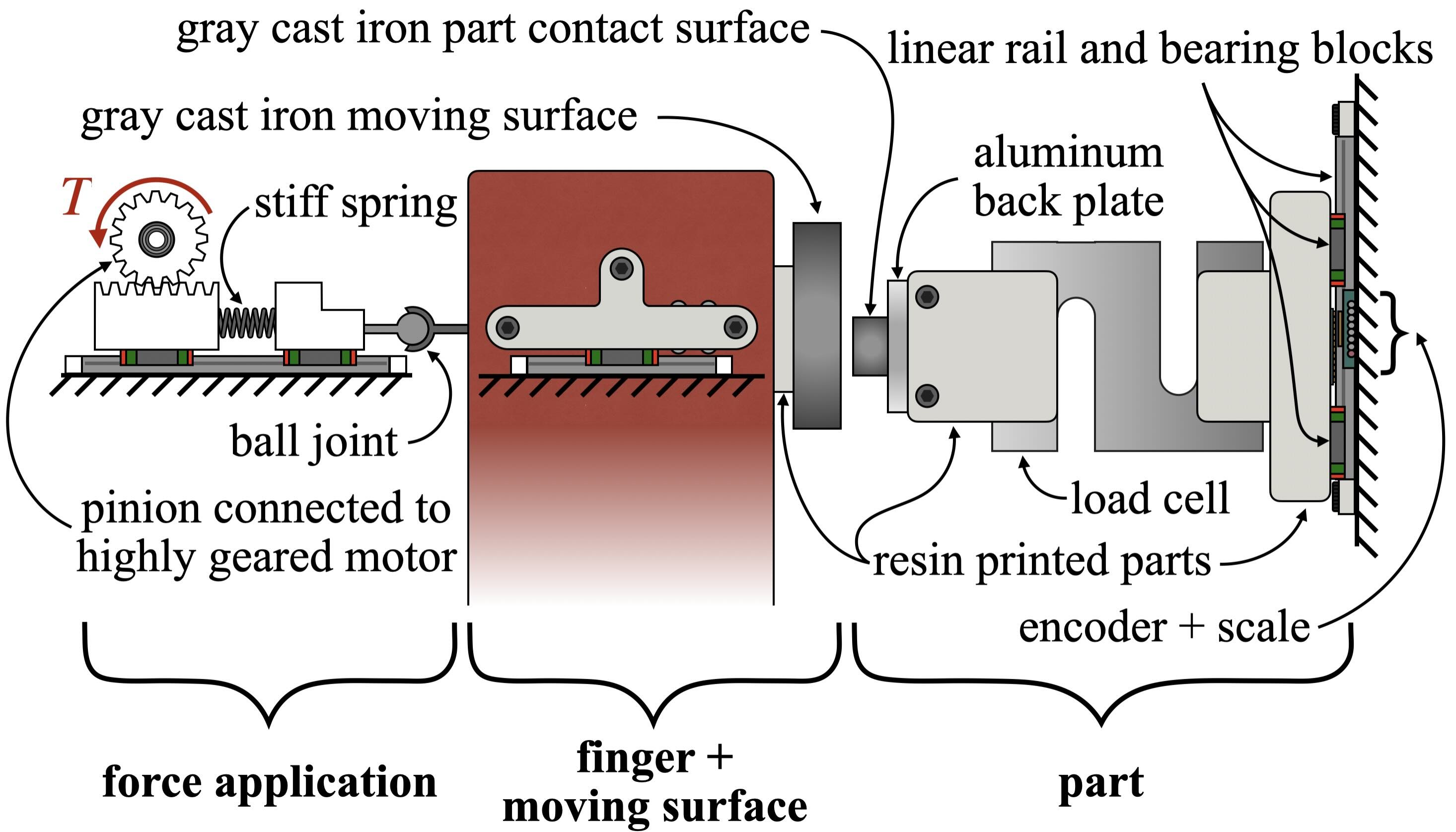}
    \caption{Main components of the experimental setup.
    Low-friction bearing blocks and rails are attached to both sides of the finger to provide smooth linear motion.
    For the experiments only, the moving surface is gray cast iron, and is pushed into the part by a highly geared motor through a stiff spring and ball joint.
    The part is constrained to move linearly along a rail via two bearing blocks, and its position is measured by a non-contact encoder.
    The part consists of a gray cast iron contact surface screwed into an aluminum backplate, a load cell, and resin components for rigidity.
    }
    \label{figure: experimental setup}
    \vspace{\shift}
\end{figure}

The finger design from Section~\ref{section: device design} with a 2-DoF gray cast iron moving surface was integrated into the experimental setup shown in Fig.~\ref{figure: experimental setup}.
The purpose of these experiments was to investigate the effect of $a_s$ and $a_{max}$ on $v_{avg}$: we measured average part velocity when the finger applied a specified normal force and vibrated with a prescribed waveform.
In the setup the finger was constrained to move linearly by low-friction bearing blocks$^{\ref{comp: vca bearing block}}$ and rails$^{\ref{comp: vca rail}}$ mounted on either side.
A highly geared motor$^{\ref{comp: experiment motor}}$ pushed the finger's moving surface into the test part with a prescribed normal force through a stiff spring and ball joint.
The test part consisted of a gray cast iron contact surface---explained below---attached to an aluminum backplate, a class C3 load cell$^{\ref{comp: loadcell}}$, and resin-printed components to increase bending stiffness of the cantilevered part. 
Vertical motion of the part was constrained by two bearing blocks$^{\ref{comp: vca bearing block}}$ sliding on a rail$^{\ref{comp: vca rail}}$. 
The gray cast iron surfaces were made parallel by aligning their centers, then pressing them together before tightening their mounting screws.

A gray cast iron on gray cast iron contact was selected because of its reported high static coefficient of friction ($\mu_s = 1.10$) and low kinetic coefficient of friction ($\mu_k = 0.15$) in Marks’ Standard Handbook for Mechanical Engineers~\cite{avallone2006marks}. 
This large difference is attractive for enabling transport at lower normal forces and for reducing wear during sliding, both useful for repeated experimental testing.
However, the measured values for our gray cast iron surfaces differed significantly, with $\mu_s = 0.44 \pm 0.04$ and $\mu_k = 0.34 \pm 0.03$ (mean $\pm$ standard deviation). 
Tracing the tabulated values back to their sources revealed that $\mu_s$ originates from Ernst and Merchant’s 1940s friction experiments on machined metals~\cite{ernst1940surface}, while $\mu_k$ originates from Morin’s 1830s work on friction~\cite{morin1832nouvelles}. 
Given the historical context and his focus on materials used in machinery, Morin most likely tested gray cast iron surfaces finished by hand scraping.
This study replicated similar conditions by progressively wet-sanding the cast iron surfaces to 1000 grit, though the moving surface was slightly roughened by tests before the formal experiments presented here.
Post-experiment review of Ernst and Merchant’s work suggests they tested lathed cast iron without further finishing, which may partly explain the discrepancy between the reported and measured values, as well as the supposed large difference in $\mu_s$ and $\mu_k$. 
Despite this, our machined and sanded gray cast iron contact surfaces proved sufficient for the presented experiments. 


The effect of $a_{max}$ on $v_{avg}$ was studied by fixing $f$ and $a_s$, and then transporting the part for different values of the $a_{max}$. 
Transport for a given waveform was also measured across progressively lower normal forces from $50$ to $30~\mathrm{N}$ in $2~\mathrm{N}$ decrements. 
The static and kinetic coefficients of friction were remeasured between each set of $a_{max}$ trials to ensure consistent contact conditions. 
To investigate the effect of $a_s$, the same procedure was followed with $f$ and $a_{max}$ fixed while $a_s$ varied. 
In this case, normal forces of $50$, $40$, and $30~\mathrm{N}$ were tested, which was deemed sufficient to capture the effect of $a_s$ based on results from the $a_{max}$ trials. 
The values of $\mu_s$ and $\mu_k$ were checked before and after all of the varied $a_s$ experiments.
The complete set of experiments in the order they were performed, as well as the measured values of $\mu_s$ and $\mu_k$, is summarized in Table~\ref{tab: experimental parameters}. 


Parameter ranges were selected based on practical constraints. 
The upper bound on $a_{max}$ was set to $20~g$, limited by the driven assembly mass of $0.524~\mathrm{kg}$ and the combined actuator force of $97.86~\mathrm{N}$. 
The maximum feasible $a_s$ was constrained by the actuator stroke of $6.10~\mathrm{mm}$, calculated from \eqref{eq: waveform amplitude}. 
The normal force range was bounded below by the minimum required to maintain sticking and above by the force beyond which no transport occurred. 
For a part mass of $0.338~\mathrm{kg}$ and $a_s = 1.0~g$, the theoretical minimum required normal force was $\approx 15~\mathrm{N}$, or $30~\mathrm{N}$ with a safety factor of two. 
The maximum normal force was selected to be less than that when~\eqref{eq: average part velocity} equals zero, i.e., when $a_{max} = a_k$. 
Transport should occur for all conditions except the case $a_{max} = 5~g$, where no transport is expected above $39~\mathrm{N}$. 

\section{RESULTS}\label{section: results}

\begin{figure}[t]
    \centering
    \includegraphics[width=0.48\textwidth]{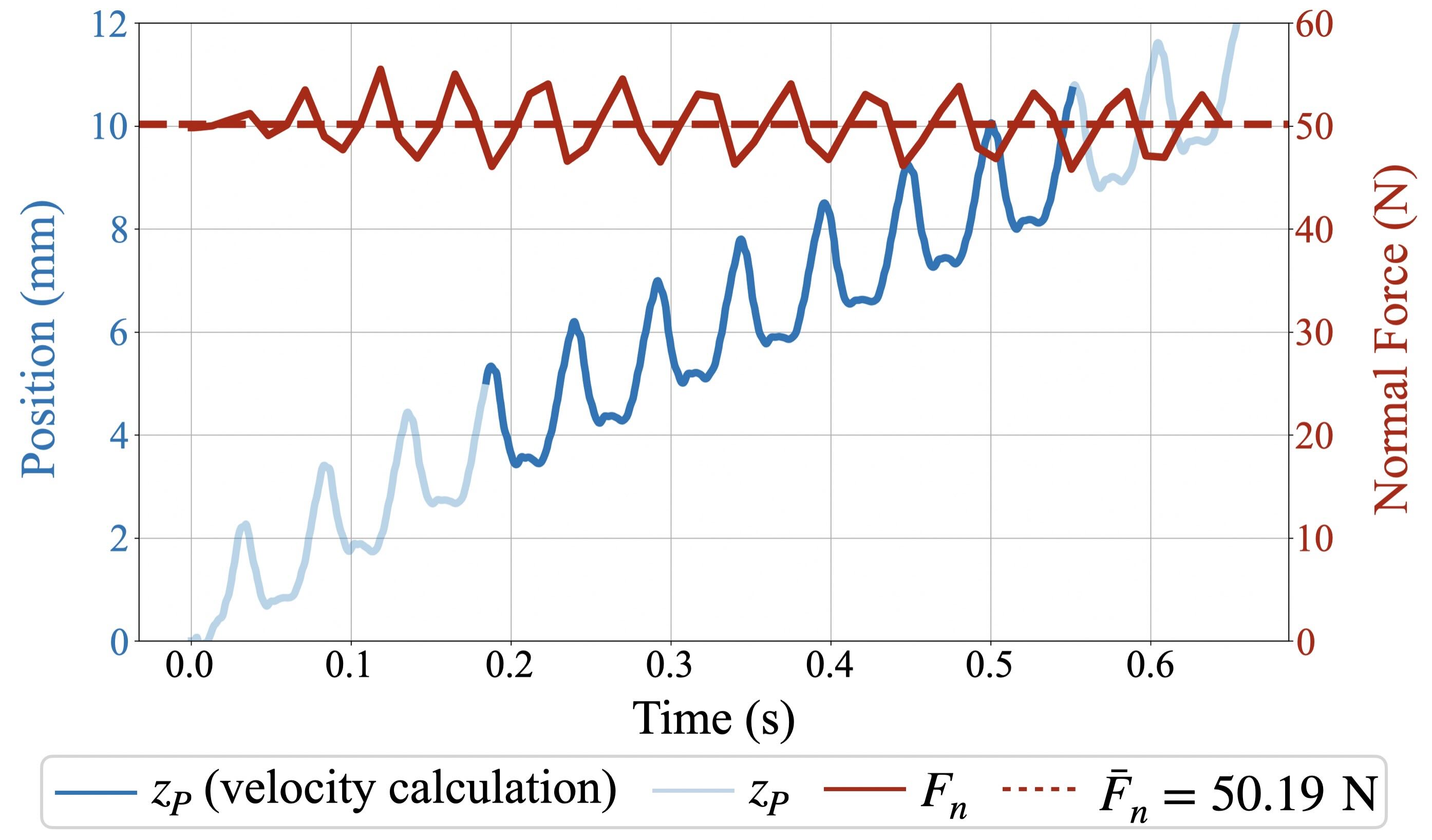}
    \caption{Sample experimental data for Experiment 3 with $F_n = 50~\mathrm{N}$.
    The recorded normal force is shown as the solid red line, with the average shown as the dashed red line.
    The part position is shown in blue, with the darkened portion of the curve representing the data used to calculate $v_{avg}$.
    }
    \label{figure: experiment plot}
    \vspace{\shift}
\end{figure}

A sample experimental plot is shown in Fig.~\ref{figure: experiment plot}. 
The solid and dashed red lines represent the recorded and average normal forces, respectively.
Since the load cell only measures axial forces but is fixed on either end, off-axis loads induce bending that causes the perceived axial load to oscillate with the part motion.
The blue lines show the part motion, with the darkened portion indicating the range where position peaks were used to calculate $v_{avg}$.
This range was selected because it is where the contact area is completely contained within the moving surface's bearing when the moving surface is centered.
In this range, bending of the moving surface from off-axis forces should be minimized, lending a closer approximation to the idealized scenario from Fig.~\ref{figure: kinematics and dynamics}.
Note that we tried other measurement ranges as well (increasing or decreasing the bounds by the same amount) and the same $a_s$ and $a_{max}$ trends discussed below persisted.

\begin{figure}[t]
    \centering
    \includegraphics[width=0.48\textwidth]{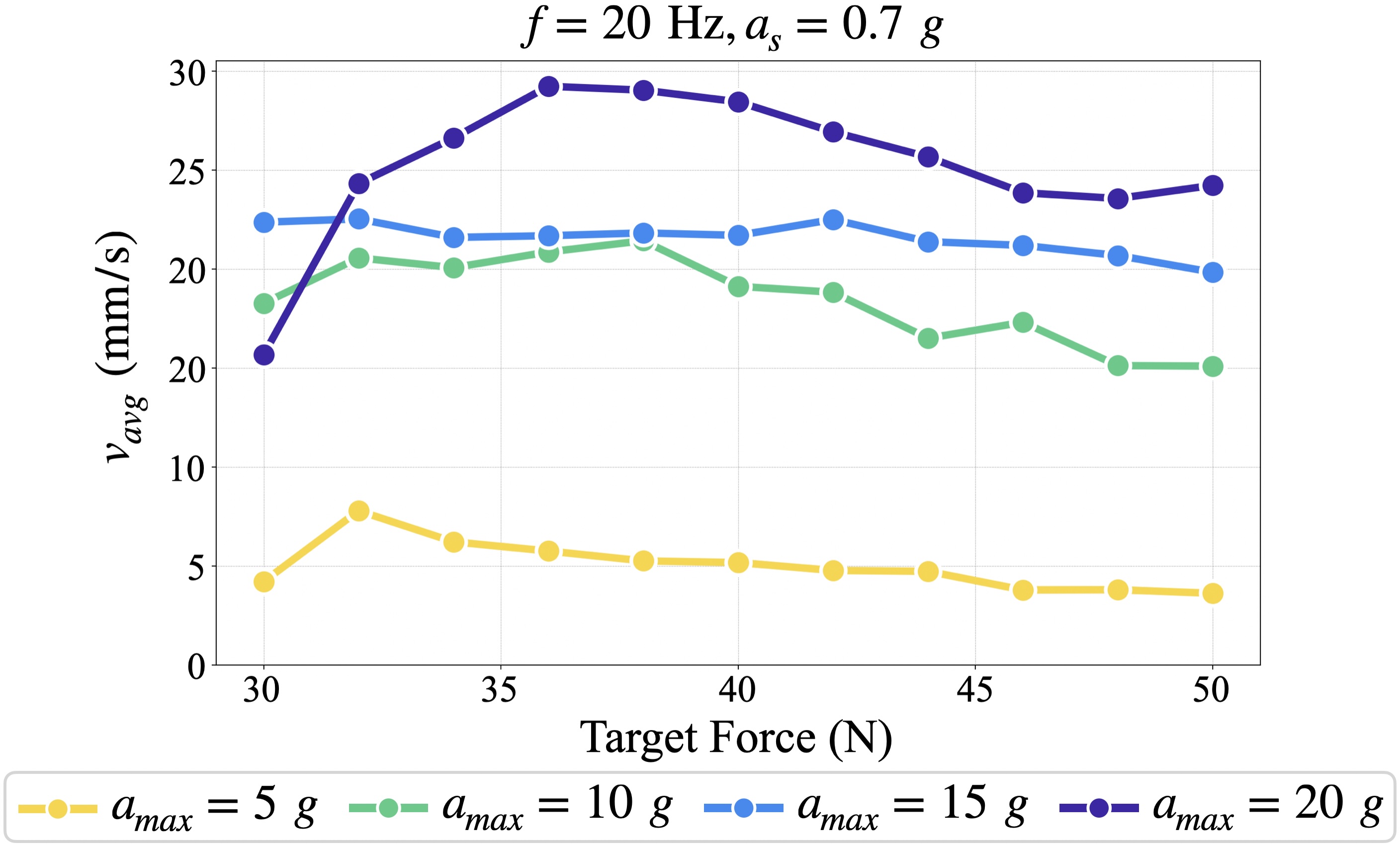}
    \caption{Average part velocity vs. target normal force for waveform parameters $f=20~\mathrm{Hz}$, $a_s = 0.7~g$, and a varying $a_{max}$ (Experiments 1 -- 4).
    Generally, the higher the maximum acceleration the higher the average part velocity, as predicted. 
    The exception is when the target force is $30~\mathrm{N}$, most likely from worsened plane-on-plane contact from the large accelerations.
    }
    \label{figure: varying amax plot}
\end{figure}
\begin{figure}[t]
    \centering
    \includegraphics[width=0.48\textwidth]{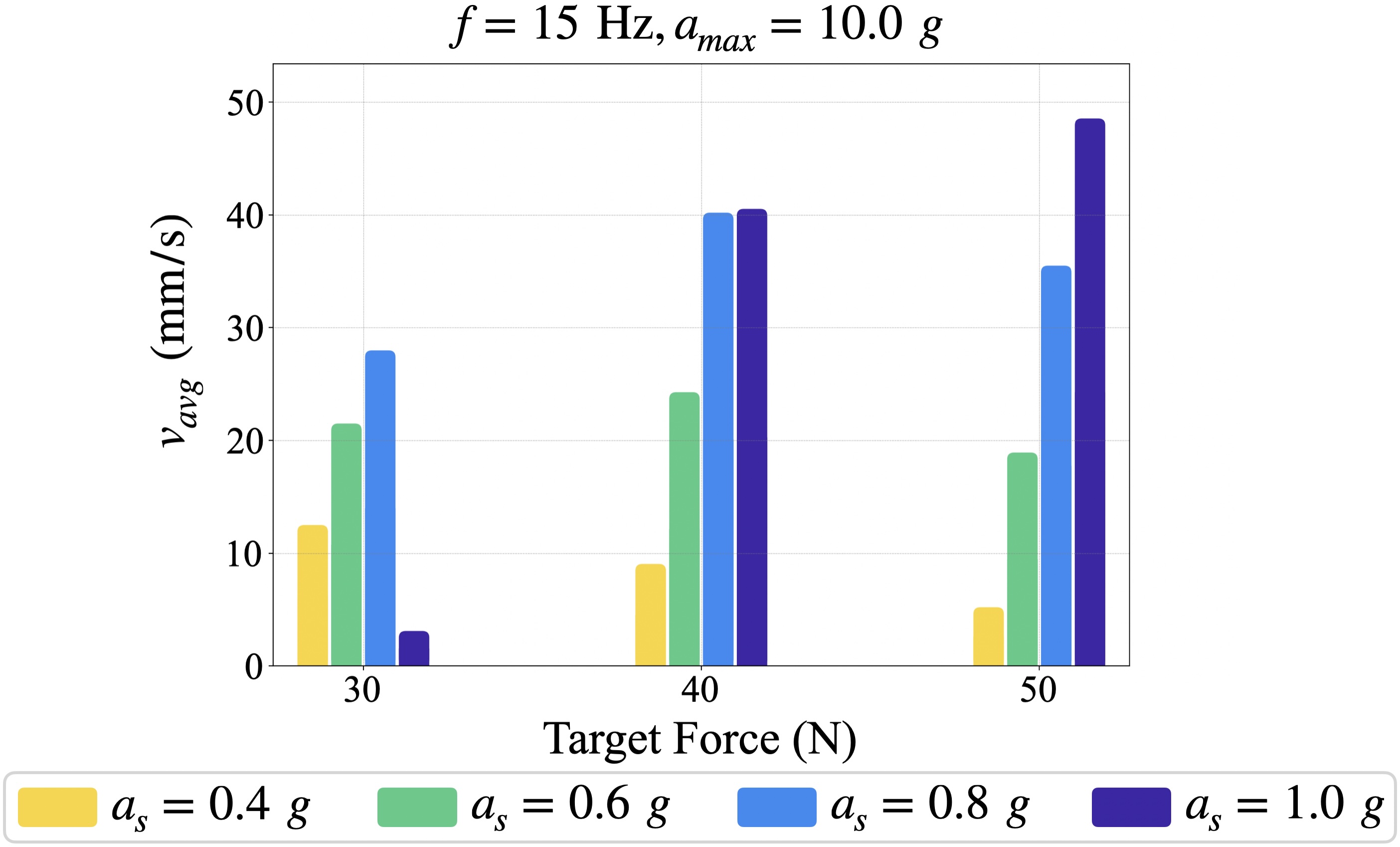}
    \caption{Average part velocity vs. target normal force for waveform parameters $f = 15~\mathrm{Hz}$, $a_{max} = 10.0~g$, and a varying $a_s$ (Experiments 5 -- 8).
    As expected, larger sticking accelerations result in increased average part velocities.
    This trend is violated when the target force is $30~\mathrm{N}$, most likely from worsened plane-on-plane contact from the large accelerations.
    }
    \label{figure: varying amin plot}
    \vspace{\shift}
\end{figure}

The results from Experiments 1 -- 8 are shown in Fig.~\ref{figure: varying amax plot} and Fig.~\ref{figure: varying amin plot}, respectively.
From the varying $a_{max}$ experiments (Fig.~\ref{figure: varying amax plot}), a higher $a_{max}$ typically leads to a higher average part velocity, $v_{avg}$, as expected from \eqref{eq: a_max partial}.
However, this trend tends to break at lower target forces---particularly for the highest maximum acceleration we tested, $a_{max} = 20~g$---while appearing distinct as the target force is increased.
We suspect this is due to several non-idealities with the setup.
Namely, the moving surface is not completely rigid and the contact surfaces are not perfectly aligned.
Higher accelerations yield stronger vibrations in the setup and greater out-of-plane motion of the moving surface.
This can worsen already imperfect plane-on-plane contact between the moving surface and the part, reducing the effective contact area and therefore the actual friction at contact.
This could explain why transport occurred with $a_{max} = 5~g$ when $F_n$ exceeded the theoretical maximum of $39~\mathrm{N}$.
Additionally, with this smaller friction cone the part was most likely spending more time slipping than sticking while the surface was driven with $a_s$, decreasing the average part velocity.
At larger normal forces the plane-on-plane contact is more strongly enforced, increasing the effective contact area and hence friction.

From Fig.~\ref{figure: varying amin plot}, a higher sticking acceleration $a_s$ generally leads to higher average part velocities.
This trend is consistent across the normal forces except for $a_s = 1.0~g$.
Just as with Experiments 1 -- 4, we believe the imperfect plane-on-plane contact and out-of-plane surface motion results in lower coefficients of friction at lower forces, especially with large accelerations.
%
Note that with $a_s = 1.0~g$ the moving surface gains roughly $23\%$ more linear momentum than when $a_s = 0.8~g$.
Given this, and the fact that the moving surface is cantilevered relative to its single 6005Z support bearing, the out-of-plane motion is increased, hence reducing the effective contact area.
%
As the normal force is increased, the plane-on-plane contact is better enforced, revealing the expected trend.

The coefficients of static and kinetic friction before / after various experiments are shown in the last two columns of Table~\ref{tab: experimental parameters}. 
No noticeable trend was observed across the experiments, indicating that surface wear from a given experiment did not significantly impact subsequent experiments.

\section{PARALLEL JAW GRIPPER}\label{section: gripper}

\begin{figure}[!t]
    \centering
    \includegraphics[width=0.48\textwidth]{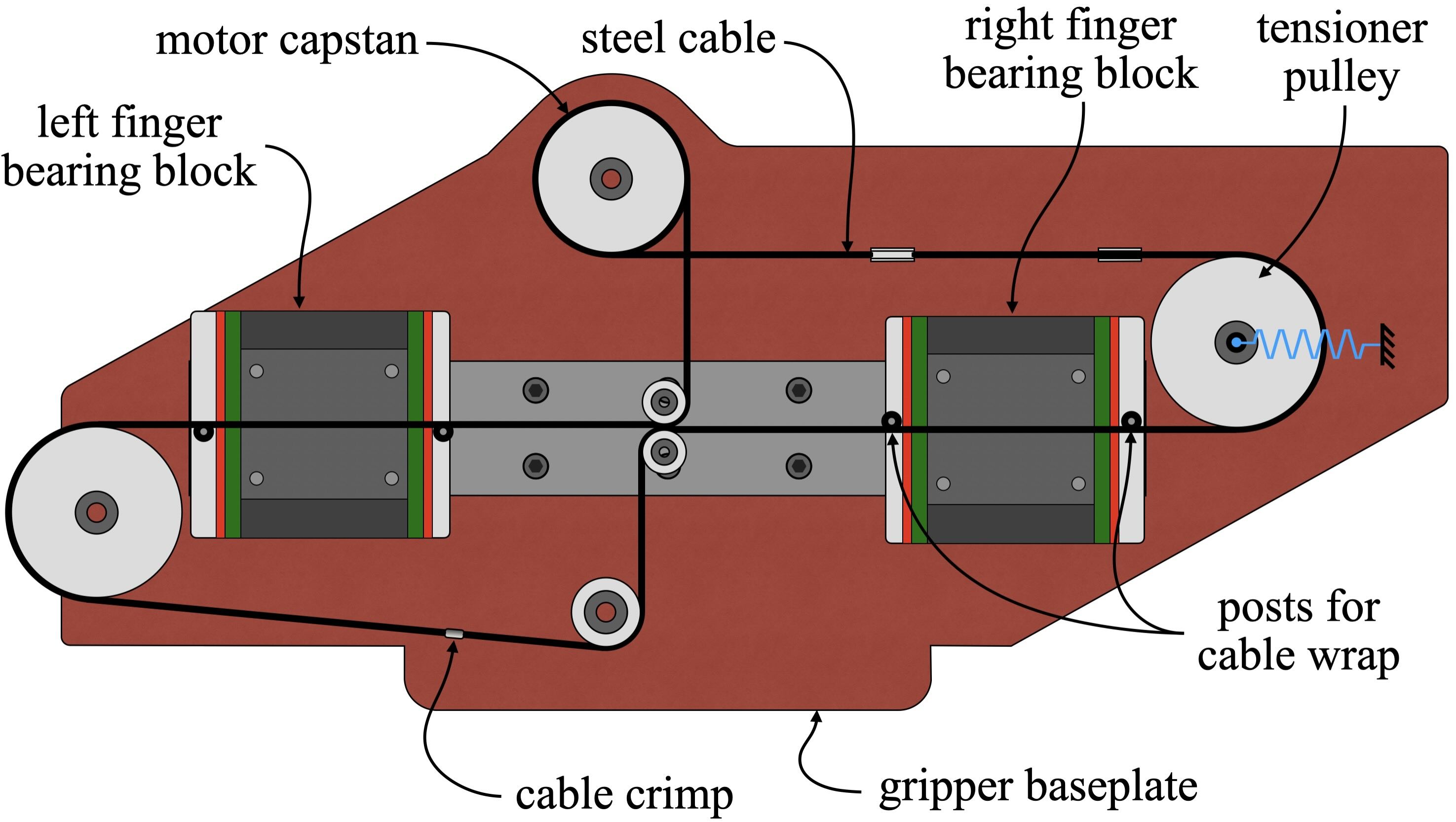}
    \caption{Cable drive used for the gripper.
    }
    \label{figure: cable drive}
    \vspace{\shift}
\end{figure}

A parallel jaw gripper using two of the 2-DoF moving surface fingers from Fig.~\ref{figure: finger design} was constructed in order to test its translational and rotational manipulation capabilities. 
{\color{black}The parts were selected to qualitatively assess the gripper's performance across a range of varying contact geometries (plane, line, point), friction properties (part surfaces such as soft fabric, metal, and plastic), and relative part compliances.}
In theory, two independently controlled 2-DoF surfaces can achieve 3-DoF manipulation: translation, in-plane rotation, and out-of-plane rotation.
Note that the gray cast iron output surface was substituted for an aluminum disk, to which a $1/16~\mathrm{inch}$-thick polyurethane sheet$^{\ref{comp: polyurethane}}$ was affixed using cyanoacrylate.
The two fingers were attached to the same $42~\mathrm{mm}$ wide rail$^{\ref{comp: gripper rail}}$ using large, low-friction bearing blocks$^{\ref{comp: gripper bearing block}}$.
This wide rail and bearing block combination was selected for its large static and dynamic moment ratings, which are necessary to counter loads at the cantilevered contact surface.
The grasping motion was driven by a single $150~\mathrm{W}$ motor$^{\ref{comp: gripping motor}}$ via the linear cable drive mechanism shown in Fig.~\ref{figure: cable drive}. 
The finger surfaces were aligned by placing a 123 block between the moving surfaces, clamping the fingers together, and progressively tightening the relevant screws.
Surface alignment with the vertical was checked with a level.

\begin{figure}[!t]
    \centering
    \includegraphics[width=0.40\textwidth]{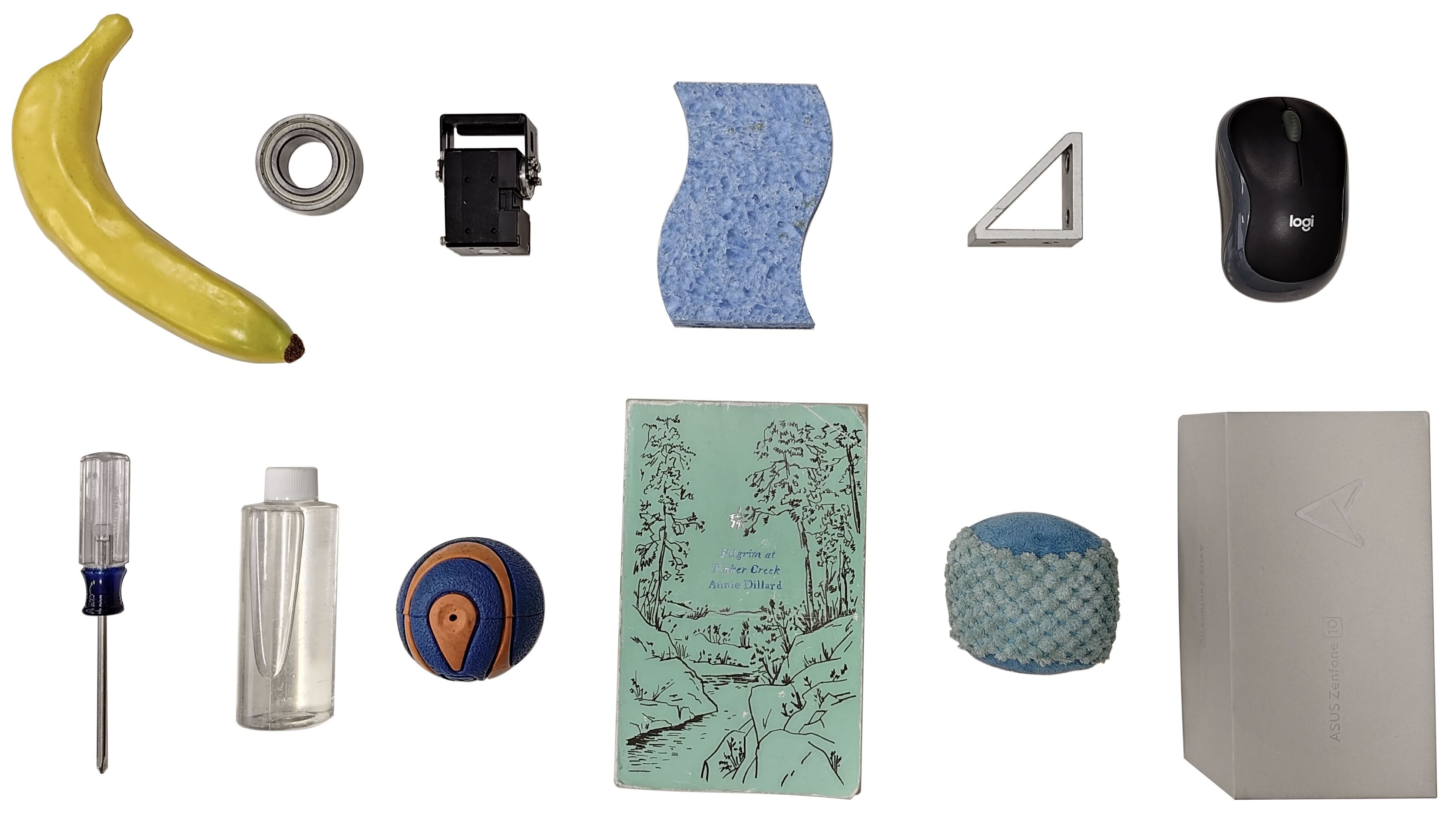}
    \caption{Transported parts.
    From left-to-right and top-to-bottom the parts are: fake banana, 6005Z bearing, Dynamixel, sponge, 80/20 bracket, mouse, screwdriver, bottle of oil, rubber dog ball, book, plush ball, and Asus Zenfone 10 box.
    }
    \label{figure: transported parts}
    \vspace{-1mm}
\end{figure}
\begin{figure}[!t]
    \centering
    \includegraphics[width=0.42\textwidth]{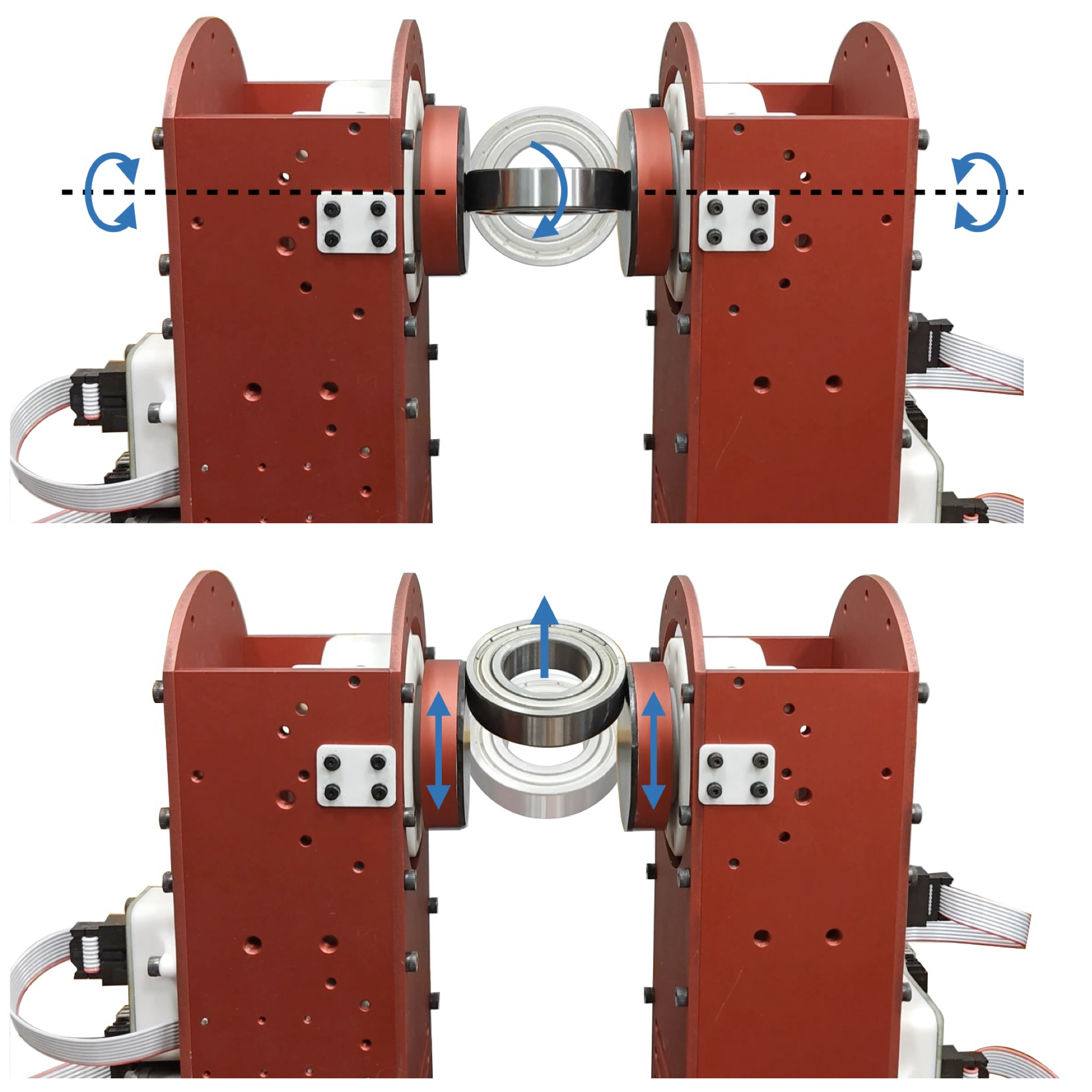}
    \caption{Example of the 6005Z bearing being manipulated.
    Blue arrows represent the motion of the moving surfaces as well as the part.
    Rotational vibration of the surfaces induces rotation of the part (top).
    Translational vibration of the surfaces results in vertical transport of the part (bottom).
    }
    \label{figure: manipulation graphic}
    \vspace{\shift}
\end{figure}

The parts manipulated in the accompanying video can be seen in Fig.~\ref{figure: transported parts} and a visualization of the manipulation using overlaid, transparent stills can be seen in Fig.~\ref{figure: manipulation graphic}.
{\color{black}For experimental reproducibility, the waveform parameters and estimated normal forces for each part are listed in Table~\ref{tab: transport_experiments}, though these are not exhaustive of those capable of manipulating a given part.}
The normal force is linearly related to the motor voltage (provided the motor is stationary), and therefore a rough initial calibration was performed to estimate the squeeze force.
Due to power supply limitations, some of the heavier parts that were manipulated required the normal force  to be applied manually (listed as \textit{hand-actuated} in Table~\ref{tab: transport_experiments}). 
For parts such as the book and the Asus Zenfone 10 box, the required normal forces to keep the parts in the grasp exceeded those achievable with the power supply.
On the other hand, the normal forces for the plush ball were near that of the friction force between the bearing block and rail and were therefore easier to apply by hand.

{\color{black}It is important to note the goal of these experiments is simply to demonstrate the controllable use of vibrating surfaces for in-hand manipulation across a diverse range of parts in a worst-case scenario for an active-surface-based manipulator, rather than to achieve precise part poses.
In fact, humans perform in-hand manipulation through a mix of sensory feedback for fine adjustment and predictive gross part motion~\cite{johansson2009coding}.
This device focuses on the latter.
}

%

\begin{table*}[ht]
\centering
\caption{Gripper Transport Experiments}
\begin{tabular}{cccccccc}
\toprule
\multirow{2}{*}{\textbf{Part}} & 
\multirow{2}{*}{\textbf{Movement Type}} & 
\multicolumn{3}{c}{\textbf{Waveform Parameters}} & 
\multirow{2}{*}{\textbf{Estimated Normal Force (N)}} & 
\multirow{2}{*}{\textbf{Bidirectional}} \\
\cmidrule(lr){3-5}
 & & $f$ (Hz) & $a_s$ ($g$) & $a_{max}$ ($g$) & & & \\
\midrule\midrule
\multirow{2}{*}{\textbf{Banana}} & \textit{Translation} & 40 & 2.0 & 20 & 2.6 & {\color{red}\xmark}\\
& \textit{Rotation} & 40 & 2.0 & 20 & 5.8 & {\color{ForestGreen}\cmark}\\
\midrule
\multirow{2}{*}{\textbf{Bearing}} & \textit{Translation} & 40 & 1.0 & 20 & 4.5 & {\color{red}\xmark}\\
& \textit{Rotation} & 40 & 3.0 & 20 & 7.0 & {\color{ForestGreen}\cmark}\\
\midrule
\multirow{2}{*}{\textbf{Dynamixel}} & \textit{Translation} & 40 & 1.0 & 20 & \textit{hand-actuated} & {\color{red}\xmark}\\
& \textit{Rotation} & 40 & 2.0 & 20 & \textit{hand-actuated} & {\color{ForestGreen}\cmark}\\
\midrule
\multirow{2}{*}{\textbf{Sponge}} & \textit{Translation} & 40 & 3.0 & 20 & 1.3 & {\color{ForestGreen}\cmark}\\
& \textit{Rotation} & 40 & 3.0 & 20 & 1.3 & {\color{ForestGreen}\cmark}\\
\midrule
\multirow{2}{*}{\textbf{80/20 Bracket}} & \textit{Translation} & 40 & 1.0 & 20 & 4.5 & {\color{red}\xmark}\\
& \textit{Rotation} & 40 & 3.0 & 20 & 4.5 & {\color{ForestGreen}\cmark}\\
\midrule
\multirow{2}{*}{\textbf{Mouse}} & \textit{Translation} & 40 & 1.0 & 20 & 4.5 & {\color{red}\xmark}\\
& \textit{Rotation} & 40 & 2.0 & 20 & 5.8 & {\color{ForestGreen}\cmark}\\
\midrule
\multirow{2}{*}{\textbf{Screwdriver}} & \textit{Translation} & 40 & 1.0 & 20 & 5.8 & {\color{red}\xmark}\\
& \textit{Rotation} & 40 & 3.0 & 20 & 12.1 & {\color{ForestGreen}\cmark}\\
\midrule
\multirow{2}{*}{\textbf{Bottle of Oil}} & \textit{Translation} & 40 & 1.0 & 20 & 3.9 & {\color{ForestGreen}\cmark}\\
& \textit{Rotation} & 40 & 3.0 & 20 & 7.7 & {\color{ForestGreen}\cmark}\\
\midrule
\multirow{2}{*}{\textbf{Rubber Dog Ball}} & \textit{Translation} & 40 & 1.0 & 20 & 2.0 & {\color{ForestGreen}\cmark}\\
& \textit{Rotation} & 40 & 2.0 & 20 & 2.0 & {\color{ForestGreen}\cmark}\\
\midrule
\multirow{2}{*}{\textbf{Book}} & \textit{Translation} & 40 & 3.0 & 20 & \textit{hand-actuated} & {\color{ForestGreen}\cmark}\\
& \textit{Rotation} & 40 & 3.0 & 20 & \textit{hand-actuated} & {\color{ForestGreen}\cmark}\\
\midrule
\multirow{2}{*}{\textbf{Plush Ball}} & \textit{Translation} & 40 & 2.0 & 20 & \textit{hand-actuated} & {\color{ForestGreen}\cmark}\\
& \textit{Rotation} & 40 & 2.0 & 20 & 3.2 & {\color{ForestGreen}\cmark}\\
\midrule
\multirow{2}{*}{\textbf{Asus Zenfone 10 Box}} & \textit{Translation} & 40 & 1.0 & 20 & 3.9 & {\color{ForestGreen}\cmark}\\
& \textit{Rotation} & 40 & 3.0 & 20 & \textit{hand-actuated} & {\color{ForestGreen}\cmark}\\
\bottomrule
\end{tabular}
\label{tab: transport_experiments}
\vspace{\shift}
\end{table*}

\section{DISCUSSION}\label{section: discussion}

\subsection{Bidirectional Translation}
Bidirectional transport proved challenging for stiffer parts as seen in Table~\ref{tab: transport_experiments}: they could move upward but had more difficulty during downward translation, even when applying a downward Quaid waveform.  
This was not strictly due to output surface misalignment, as upward-only transport could persist even when the surfaces were intentionally angled downward, likely a consequence of the cantilevered finger design.  
Slow-motion video revealed that the momentum of the moving surface induced a periodic out-of-plane tilt, exploiting play between the inner and outer races of the 6005Z bearing that supports the moving surface.  
These out-of-plane motions generate reaction forces that, especially for stiffer parts, transmit across the coupled fingers and can disrupt contact. 
The effects of geometric bias from the cantilevered fingers and unintended out-of-plane surface motion likely led to the experimentally observed ``upward-only'' transport for stiffer parts.
%
%
%
%
Several design improvements could enable consistent bidirectional rigid-part translation: using non-cantilevered fingers, replacing the single ball bearing with dual angular contact or tapered roller bearings to minimize tilt, decoupling the finger open / close motion, and reducing the moving mass.
The goal of the last design improvement is to minimize the effect of any inertial forces along the gripper closing / opening direction due to misalignment between the output surface rails and the vertical.

To the authors' surprise, many compliant parts were easily translated bidirectionally. 
This compliance appears to have the effect of absorbing out-of-plane motion, hence minimizing force transmission between the fingers and enabling consistent contact with the moving surface.
In our previous work~\cite{yako2024vertical}, compliance is inherent to the gripper design; the moving surface is connected to the gripper through flexures that can absorb reaction forces from unintended motion.
Additionally, part compliance enables accommodation of the upward geometric bias from the cantilevered fingers through deformation rather than the upward rigid-body motion observed with stiffer parts.
Where to introduce compliance in the design presented here is still a key open question, but may not be necessary if strictly handling compliant parts or if the stated mechanical design changes improve the behavior.

\subsection{Bidirectional Rotation}
Out-of-plane rotation was somewhat successful, but only for select parts and typically required external support to prevent parts from falling or rotating out of the grasp.
Just as with translation, out-of-plane motion of the gripper surfaces most likely impeded clean rotation of the part, so minimizing it could lead to more consistent bidirectional translation and out-of-plane rotation.
In-plane rotation worked well and was agnostic to part stiffness.
We also found that generally in-plane rotation required larger squeeze forces compared with translation.
Note that in most cases the part will rotate at some radius relative to the center of the moving surfaces. 
The translational capability of the device can be used to control this radius, as well as draw in parts that may be rotating out of the grasp.
It was also observed that the structural vibrations during in-plane rotation were typically smaller compared with translation.
We hypothesize this is due to the decreased moving mass during rotation, since the bearing holder and outer bearing race are stationary.
%

\subsection{Experimental Results and Practical Considerations}
The waveform parameters and squeeze forces reported in Table~\ref{tab: transport_experiments} did not require fine tuning to achieve transport.  
In practice, transport was easy to obtain and occurred over a broad range of parameters and forces.  
A waveform frequency of $40~\mathrm{Hz}$ was used because it provided a balance between transport speed and reaction time for a human operator selecting transport type (translation / rotation) and direction (up / down or clockwise / counterclockwise) in real-time.
The experiments highlight several practical guidelines to maximize the speed of part transport, which were also applied in the demos in the accompanying video.  
First, $a_{max}$ should be set to $a_{system,max}$, estimated as $20~g$ for our device and used across all parts. 
Performance did not seem sensitive to overestimating $a_{system,max}$, so only a rough value is needed.  
The sticking acceleration $a_s$ should also be maximized, unless excessive vibrations or out-of-plane motion arise, in which case it should be reduced.  
Previously mentioned mechanical design improvements could allow for larger values of $a_s$ for a given $f$.
If no transport occurs with $a_s$ maximized at a given frequency (i.e., $a_s : A(a_s, a_{max}, T) = A_{max}$), the frequency should be increased to allow a larger feasible $a_s$. 
The usable frequency range can be estimated from the VCA datasheet, moving mass, and desired amplitude.


\section{CONCLUSIONS AND FUTURE WORK}\label{section: conclusion and future work}

Our research demonstrated a system capable of achieving vibratory-based translation and rotation of parts held in a grasp.
We discussed the dynamical model and showed how average part velocity is affected by the alternating sticking and slipping accelerations in the Quaid waveform~\cite{quaid1999feeder}.
Namely, the sticking and slipping accelerations should be maximized in order to maximize part velocity.
We then developed a system that can effectively track the desired stick-slip surface waveform, and used this to successfully reproduce the trends predicted by the theory.
The experiments also revealed some practical considerations.
Specifically, sufficient normal force is required to maintain contact in the presence of structural oscillations and plane-on-plane misalignment.
The system used in the experiments was integrated into a parallel jaw gripper, where each finger had a 2-DoF moving surface.
A variety of parts were bidirectionally translated and rotated.
Stiffer parts could be transported upwards against gravity, but struggled when moving downwards, most likely due to the cantilevered finger design and out-of-plane motions resulting from play in the bearing.
On the other hand, compliant parts were easily bidirectionally translated and rotated, as they could filter out these effects.
The gripper also demonstrated an ability to perform out-of-plane rotation of cylindrical parts, given they were sufficiently externally supported.
This work takes a step toward vibratory-based robotic in-hand manipulation.
We illustrated its application to 2-DoF in-hand manipulation, and with careful mechanical design and precise force control this approach can be extended to higher-DoF manipulation.
{\color{black}We also surmise that such a system could be used for closed-loop part manipulation, where an external camera provides feedback to determine the applied normal force and vibration waveform to manipulate a part to a given pose.}

\addtolength{\textheight}{-18cm}   




\bibliographystyle{ieeetr}
\bibliography{root}

@book{yuan2022robot,
  title={Robot in-hand manipulation using Roller Graspers},
  author={Yuan, Shenli},
  year={2022},
  publisher={Stanford University}
}

@inproceedings{yako2024vertical,
  title={Vertical Vibratory Transport of Grasped Parts Using Impacts},
  author={Yako, Connor L and Nowak, J{\'e}r{\^o}me and Yuan, Shenli and Salisbury, Kenneth},
  booktitle={2024 IEEE International Conference on Robotics and Automation (ICRA)},
  pages={1950--1956},
  year={2024},
  organization={IEEE}
}

@book{avallone2006marks,
  title={Marks' Standard Handbook for Mechanical Engineers. 10},
  author={Avallone, Eugene and Baumeister, I and Sadegh, Ali},
  year={2006},
  publisher={Citeseer}
}

@inproceedings{ernst1940surface,
  title={Surface friction of clean metals: a basic factor in the metal cutting process},
  author={Ernst, HEME and Merchant, M EUGENE},
  booktitle={Proc. MIT Conf., on Friction and Surface Finish},
  pages={76--101},
  year={1940},
  organization={The MIT Press Cambridge, MA}
}

@book{morin1832nouvelles,
  title={Nouvelles experiences sur le frottement},
  author={Morin, Arthur},
  volume={1},
  year={1832},
  publisher={Bachelier}
}

@article{nahum2022robotic,
  title={Robotic manipulation of thin objects within off-the-shelf parallel grippers with a vibration finger},
  author={Nahum, Noam and Sintov, Avishai},
  journal={Mechanism and Machine Theory},
  volume={177},
  pages={105032},
  year={2022},
  publisher={Elsevier}
}

@article{binyamin2024vibration,
  title={Vibration-based Full State In-Hand Manipulation of Thin Objects},
  author={Binyamin, Oron and Shapira, Guy and Nahum, Noam and Sintov, Avishai},
  journal={arXiv preprint arXiv:2412.14899},
  year={2024}
}

@INPROCEEDINGS{reznik1998coulomb,
  author={Reznik, D. and Canny, J.},
  booktitle={Proceedings. 1998 IEEE International Conference on Robotics and Automation (Cat. No.98CH36146)}, 
  title={The Coulomb pump: a novel parts feeding method using a horizontally-vibrating surface}, 
  year={1998},
  volume={1},
  number={},
  pages={869-874 vol.1},
  doi={10.1109/ROBOT.1998.677094}
}

@INPROCEEDINGS{quaid1999feeder,
  author={Quaid, A.E.},
  booktitle={Proceedings 1999 IEEE International Conference on Robotics and Automation (Cat. No.99CH36288C)}, 
  title={A miniature mobile parts feeder: operating principles and simulation results}, 
  year={1999},
  volume={3},
  number={},
  pages={2221-2226 vol.3},
  doi={10.1109/ROBOT.1999.770436}
}

@ARTICLE{umbanhowar2008optimal,
  author={Umbanhowar, Paul and Lynch, Kevin M.},
  journal={IEEE Transactions on Automation Science and Engineering}, 
  title={Optimal Vibratory Stick-Slip Transport}, 
  year={2008},
  volume={5},
  number={3},
  pages={537-544},
  doi={10.1109/TASE.2008.917021}
}

@inproceedings{cai2023hand,
  title={In-hand manipulation in power grasp: Design of an adaptive robot hand with active surfaces},
  author={Cai, Yilin and Yuan, Shenli},
  booktitle={2023 IEEE International Conference on Robotics and Automation (ICRA)},
  pages={10296--10302},
  year={2023},
  organization={IEEE}
}

@ARTICLE{gomezdegabriel2021adaptive,
  author={Gómez-de-Gabriel, Jesús M. and Wurdemann, Helge A.},
  journal={IEEE Robotics and Automation Letters}, 
  title={Adaptive Underactuated Finger With Active Rolling Surface}, 
  year={2021},
  volume={6},
  number={4},
  pages={8253-8260},
  doi={10.1109/LRA.2021.3105729}
}

@INPROCEEDINGS{tincani2012velvet,
  author={Tincani, Vinicio and Catalano, Manuel G. and Farnioli, Edoardo and Garabini, Manolo and Grioli, Giorgio and Fantoni, Gualtiero and Bicchi, Antonio},
  booktitle={2012 IEEE/RSJ International Conference on Intelligent Robots and Systems}, 
  title={Velvet fingers: A dexterous gripper with active surfaces}, 
  year={2012},
  volume={},
  number={},
  pages={1257-1263},
  doi={10.1109/IROS.2012.6385939}
}

@article{datseris1985principles,
    author={Datseris, P. and Palm, W.},
    title={Principles on the Development of Mechanical Hands Which Can Manipulate Objects by Means of Active Control},
    journal={Journal of Mechanisms, Transmissions, and Automation in Design},
    volume={107},
    number={2},
    pages={148-156},
    year={1985},
    month={06},
    issn={0738-0666},
    doi={10.1115/1.3258703},
    url={https://doi.org/10.1115/1.3258703},
    eprint = {https://asmedigitalcollection.asme.org/mechanicaldesign/article-pdf/107/2/148/5935929/148_1.pdf},
}

@inproceedings{ma2016manipulation,
  title={In-hand manipulation primitives for a minimal, underactuated gripper with active surfaces},
  author={Ma, Raymond R. and Dollar, Aaron M.},
  booktitle={International Design Engineering Technical Conferences and Computers and Information in Engineering Conference},
  volume={50152},
  pages={V05AT07A072},
  year={2016},
  organization={American Society of Mechanical Engineers}
}

@ARTICLE{xie2024belted,
  author={Xie, Gregory and Holladay, Rachel and Chin, Lillian and Rus, Daniela},
  journal={IEEE Robotics and Automation Letters}, 
  title={In-Hand Manipulation With a Simple Belted Parallel-Jaw Gripper}, 
  year={2024},
  volume={9},
  number={2},
  pages={1334-1341},
  doi={10.1109/LRA.2023.3346750}
}

@article{johansson2009coding,
  title={Coding and use of tactile signals from the fingertips in object manipulation tasks},
  author={Johansson, Roland S and Flanagan, J Randall},
  journal={Nature Reviews Neuroscience},
  volume={10},
  number={5},
  pages={345--359},
  year={2009},
  publisher={Nature Publishing Group UK London}
}

@Article{keek2021design,
  AUTHOR = {Keek, Joe Siang and Loh, Ser Lee and Chong, Shin Horng},
  TITLE = {Design and Control System Setup of an E-Pattern Omniwheeled Cellular Conveyor},
  JOURNAL = {Machines},
  VOLUME = {9},
  YEAR = {2021},
  NUMBER = {2},
  ARTICLE-NUMBER = {43},
  URL = {https://www.mdpi.com/2075-1702/9/2/43},
  ISSN = {2075-1702},
  DOI = {10.3390/machines9020043}
}

\end{document}